\documentclass[11pt,letterpaper]{article}
\usepackage{emnlp2016}
\usepackage{times}
\usepackage{latexsym}
\usepackage{graphicx}
\usepackage{float}

\emnlpfinalcopy

\def\emnlppaperid{***}

\graphicspath{ {images/} }

\title{Unified Framework for Quantification}

\author{Aykut Firat\\
	    Crimson Hexagon\\
	    Boston, MA, USA\\
	    {\tt aykut@crimsonhexagon.com}
}

\date{}

\begin{document}

\maketitle

\begin{abstract}
Quantification is the machine learning task of estimating test-data class proportions that are not necessarily similar to those in training. Apart from its intrinsic value as an aggregate statistic, quantification output can also be used to optimize classifier probabilities, thereby increasing classification accuracy. We unify major quantification approaches under a constrained multi-variate regression framework, and use mathematical programming to estimate class proportions for different loss functions. With this modeling approach, we extend existing binary-only quantification approaches to multi-class settings as well. We empirically verify our unified framework by experimenting with several multi-class datasets including the Stanford Sentiment Treebank and CIFAR-10.
\end{abstract}

\section{Introduction}

In business applications such as opinion monitoring on social media, accurately predicting the proportion of each opinion class is more important than predicting each individual opinion accurately.  Understanding the customer intent to watch a movie before and after the release of a trailer, for example, requires accurate quantification of the class proportions that may exhibit high variation day to day and be completely different than training-time proportions. Naively aggregating standard classifier probability outputs, in general, produces severely biased estimates of class proportions whenever test and training class distributions significantly differ, precisely when something interesting happens. In that case, probability outputs are not optimal for classification either, because the assumed prior class probabilities, namely the training proportions, no longer match the actual priors corresponding to test class proportions.

The goal of quantification is to yield accurate estimates of the test class proportions regardless of the training class proportions. These estimates can then be used to adjust classifier probability outputs. The benefits are more pronounced when classes are less separable, and test and training distributions are more divergent. Obviously, a perfect classifier is also a perfect quantifier. Despite all the advances in machine learning, however, we do not often have access to perfect classifiers \cite{hand_classifier_2006}, especially when training data is scarce. 

Our main contribution here is unifying major quantification approaches mentioned in the literature under a single framework. In this unified view, all the major quantification approaches share a multi-variate constrained regression model, and differ only in their choice of feature transformation and loss functions. This unification not only offers conceptual clarity and simplicity around the topic of quantification, but has practical benefits as well. By outlining the mathematical programming solutions for the constrained regression model for different loss functions, we extend existing binary-only quantification approaches to multi-class settings. Existing major quantification approaches will be explained next, before we detail the constrained regression framework.

\section{Related Work}

Although its history goes back to at least 1966 \cite{gart_comparison_1966}, quantification, as a name, first occurred in a series of papers by Forman \cite{forman_counting_2005,forman_quantifying_2007,forman_quantifying_2008}.  Forman identified the importance of the problem at HP, as they needed to monitor the fast diverging prevalence of support issues over time by analyzing support log data. 

In a binary setting, the classical quantification method is to first estimate the true positive rate ($tpr$) and false positive rate ($fpr$) using  a classifier with cross-validation. Then by using an adjustment formula on naive proportion prediction $p^{'}$, the adjusted class proportion $p$ is estimated as:  

\begin{center}$p = max(min(\frac{p^{'}- fpr}{tpr-fpr},1),0)$ \end{center}

Forman introduces variations of this idea by using different cut-off thresholds in calculating $p^{'}, tpr$ and $fpr$. The threshold determines the class decision boundary, and directly affects the class assignments; hence $p^{'}, tpr$ and $fpr$. The following methods based on threshold choices are described in his papers: 
\begin{enumerate} 
\item The standard threshold of $0.5$ is called adjusted count (AC)
\item The threshold that maximizes the $tpr-fpr$ is called Max
\item The threshold where $tpr = 1- fpr$ is called X
\item The threshold where $tpr = 0.5$ is called T50
\item Using every quantized threshold first and then taking the median of the estimates is called Median Sweep (MS)
\end{enumerate}

While these thresholds are empirically chosen, Friedman uses the optimum threshold that minimizes the variance of proportion estimates \cite{friedman_class_2014}. Friedman's optimum threshold is dynamic and for a given class corresponds to its training proportion. 

Forman also introduces a mixture model, where he seeks to combine classifier score distributions per class obtained via cross-validation  with test class proportions to obtain the test classifier score distribution. Test class proportions that minimize the difference (using a loss function called PP-Area that corresponds to minimizing L1-norm) between the predicted test CDF and actual CDF then becomes his solution. A similar idea is used by \cite{gonzalez-castro_class_2013} in their HDy method with basically two differences: they replace Forman's PP-Area metric with Hellinger distance, and use probability mass function (PMF) instead of the CDF to characterize the distributions. The method used by \cite{bella_quantification_2010} can also be viewed as a mixture model. Instead of using a PMF or CDF, they use the mean value of the probability distributions in the mixture model. Friedman, on the other hand, utilizes a mixture model with PMF  in setting up a diagnostic test for concept drift, i.e. whether the distribution of features per class is still the same in training and test, but does not utilize it in predicting the proportions.

The quantification approach by \cite{hopkins_method_2010}  is different from the above-mentioned methods in that it does not require a classifier to come up with the class proportion estimates. Yet this method can be seen as quite similar to the mixture model approach mentioned above with the exception that instead of classifier probabilities, feature distributions are used directly with a feature sampling strategy. Feature distribution is modeled as a univariate multinomial distribution with $2^N$ possible outcomes ($N$=sampled feature size), each of which is a possible feature profile. The HDx method outlined in \cite{gonzalez-castro_class_2013} similarly works directly with features without a classifier.

A completely different approach is pursued in \cite{esuli_optimizing_2015} and \cite{barranquero_quantification-oriented_2015}. They replace the standard loss function used in training with one that targets proportion estimation directly. These methods do not use any post-adjustment and directly use the classifier output for quantification. A recent study, \cite{tasche_does_2016}, claims that  ``quantification without adjustments'' methods work only in the degenerate case, i.e. when the class distributions in training and test are the same or nearly the same. We exclude these type of methods from our unification framework.

All of these approaches, with the exception of Friedman, and Hopkins \& King, limit their focus to binary classification, and leave the multi-class case for future work. Both Friedman, and Hopkins \& King introduce standard constrained least squares as their solution methodology for multi-class settings. Friedman, in addition, demonstrates the infeasibility of ``one vs. rest" approaches in extending quantification from binary to multi-class settings.

\section{Unifying Quantification Methods}

We now show that existing threshold and mixture-model based quantification approaches can be modeled as a constrained regression problem by identfying their feature transformation and loss functions. Because the framework is multi-variate, existing binary-only approaches are automatically extended to the multi-class setting under this modeling. 

We assume that the distribution of features given a class is identical in test and training data. Any quantification task that does not largely satisfy this fundamental assumption is beyond the scope of this paper. 

\subsection{Notation}

We will be adopting a similar notation used in \cite{friedman_class_2014}: 
\begin{description}
 \item $T=\{y_i,x_i\}_1^{N_T}$  is the labeled training data of size $N_T=\sum_{i=1}^K N_i$ with $K$ classes, where $N_i$ is the number of training examples for class $i$, and $y \in \{c_1,c_2,...,c_K\}$  is the class label variable, and $x$ is the feature vector variable of size $V$; $y_i$ and $x_i$ refer to actual values of these variables in the data.

\item $F = \{x_i\}_1^{N_F}$, is the future unlabeled data of size $N_F$

\item $I(z)$ is an indicator function:  $1$ when $z$ is $true$, $0$ otherwise

\item $\pi_{Tk} = \frac{1}{N_T}\sum_{i \in T} I(y_i = c_k)$, proportion of class $k$ in training

\item $\hat{\pi}_{Fk}$ is the estimated proportion of class $k$  in unlabeled future data,$\mathbf {\hat{\pi}}_F = ( \hat{\pi}_{F1}, \hat{\pi}_{F2},...,\hat{\pi}_{FK})$, a column vector

\item $\hat{y}_i =$ $argmax_{k_1^K}$ $\hat{P}_{Tk}(x_i)$ , where $\hat{P}_{Tk}(x)$  is the machine learning estimate of $P_{Tk}(x)=Pr(y=c_k|x)$ 

\item $f_l(x) $ is a feature transformation function to be defined later depending on the quantification approach, $l=1..L, L \geq K$

\item $\hat{f}_{l}^F = \frac{1}{N_F}\sum_{i \in F} f_l(x_i)$

\item $\hat{f}_{lk} = \frac{1}{N_k}\sum_{i \in T_k} f_l(x_i) $

\item $\mathbf X$  is a matrix of size $L\times K$ where $\mathbf X_{{lk}}=\hat{f}_{lk}$

\item $\mathbf y$  is a vector of length $L$ where $\mathbf y_{{l}}=\hat{f}^F_{l} $

\item $\mathbf \epsilon$  is an error vector of length $L$  

\item $bin(p)$ is one-hot encoding function that produces the vector output of size B (number of bins) with a value 1 for the bin p belongs, and zero elsewhere

\item $cumsum(v)$ is a function that outputs a vector z where $z_i=\sum_{j=1}^i v$, $v$ and $z$ are vectors of size $n, i=1..n$

\item $ntuple(v)$ produces a sparse vector of size $2^{|v|}$ with a value 1 for the cell v belongs, and zero elsewhere   

\item $sample(x,n)$ takes a random subset of size n from x

\item Naive estimate $\hat{\pi}_{Fk}^{naive} = \frac{1}{N_F}\sum_{i \in F} \hat{P}_{Tk}(x_i)$

\end{description}

\subsection{Constrained Regression}

The general constrained regression model for quantification is:
\begin{description}

\item $\mathbf y  = \mathbf X \mathbf{\hat{\pi}}_F + \mathbf \epsilon $

\item s.t.
 
\item $ \mathbf{\hat{\pi}}_F>= 0$  and  $\sum\mathbf{\hat{\pi}}_F=1$
	
\end{description} 

Existing quantification approaches differ from each other simply by adopting either a different feature transformation function, $f_l(x)$, or a different loss function for the regression. 

Hopkins \& King use a feature transformation function that directly targets approximating the distribution of features via repeated random sampling. Roughly, they produce vectors, by first sampling \textit{n} features, tabulating all existing permutations and normalizing them to derive a distribution. Gonz\'alez-Castro in their HDx method, on the other hand, derive a distribution for each feature separately. In these two methods maximum value of the subscript $l$ refers to the number of repeated samples, and number of features respectively.

The remaining approaches use a feature transformation function that operates on the probability output, $Pr(y=c_l|x)$, ($L=K$) of a base classifier.  Because the probability output is only affected by the same variables, the distribution of probability output given a class is also the same in test and training data by an extension of the fundamental assumption (see \cite{friedman_class_2014} for an illustration).  
\begin{table*}[t]
\begin{center}
\def\arraystretch{1.5}
\begin{tabular}{ |c|c|c|c|} 
 \hline
 \bf Approach & \bf {$f_l(x)$}  & \bf Loss function & \bf Multi-class  \\ 
  \hline
 Hopkins \& King (VA) & $ntuple(x[s_l]), s_l=sample(1..V,n)$ &L2 & Yes \\ 
 
Gonz\'alez-Castro et al. (HDx) & $ntuple(x[l]), l \in \{1..V\}$ & Hellinger & No\\
 
Friedman (FM) & $I(\hat{P}_l^T(x) \geq \pi_l^T)$  & L2 & Yes\\ 
 
Adj. Count (AC) &  $I(\hat{P}_l^T(x) \geq \hat{P}_{\hat{y}}^T(x))$ & - & No\\ 
 
Forman (MS) & $I(\hat{P}_l^T(x) \geq \alpha),  \forall \alpha$ & - & No\\

Bella et. al. (Prob) & $\hat{P}_l^T(x)$ & - & No\\
 
Forman (MM) & $cumsum(bin(\hat{P}_l^T(x)))$ & L1 & No\\

Gonz\'alez-Castro et al. (HDy) & $bin(\hat{P}_l^T(x))$ & Hellinger & No\\

 \hline
\end{tabular}
\end{center}
\caption{Unification: choice of feature transformation function $f_l(x)$ in different quantification techniques. The acronyms in parentheses in the Approach column will be used from now on to refer to each method. In the loss-function column: L2=least-squares, L1=least-absolute-deviation. For some methods loss-function is not explicitly specified in the binary version, they are shown with a - in the table. }
\end{table*}

The specific $f_l(x)$ choice for each quantification approach, derived simply from definitions, is shown in Table 1 along with abbreviations we will be using in the rest of the paper.

Note that, we abuse the notation slightly, when $f_l(x)$ returns a vector output as in the approaches that use the bin function. In those cases the $\mathbf X$ and  $\mathbf y$ should be flattened to change their row size from $L$ to $L\times B$. We also consider a subset of Forman's methods that have clear interpretations in multi-class settings. The T50, X and Max methods, for example, are binary specific and need a new definition for multi-class; therefore they are excluded.

It is also important to note that approaches using a base classifier need to use cross-validation to estimate the $f_l(x)$. Repeating the cross-validation multiple times, and averaging the results increases the stability of the final estimate. As in Hopkins \& King, the regression can be performed until a convergence criterion is reached. It is also possible to stack the data from many trials and solve one final regression problem.

\subsection{Solution for Different Loss Functions}

The constrained regression problem in Section 3.2 can be solved using different loss functions. In the literature three different loss functions, namely least-squares, least-absolute deviation and Hellinger divergence, have been used, so we will outline the solution methodology for these three. 

\subsubsection{Least-Squares}
The least squares problem can be formulated as follows:

\begin{description}
\item min $ (\mathbf y  - \mathbf X \mathbf{\hat{\pi}}_F)^T(\mathbf y  - \mathbf X \mathbf{\hat{\pi}}_F)$

\item s.t.  $ \mathbf{\hat{\pi}}_F>= 0$  and  $\sum\mathbf{\hat{\pi}}_F=1$

\end{description}

The objective function can be expanded as follows:
\begin{description}
\item $ \mathbf{\hat{\pi}}_F^{T}\mathbf X^T\mathbf X\mathbf{\hat{\pi}}_F-2\mathbf{\hat{\pi}}_F^{T}\mathbf X^T\mathbf y+\mathbf y^T\mathbf y$

\item Let $D = 2 \mathbf X^T\mathbf X$ and $d=-2\mathbf X^T\mathbf y$ and ignore the constant $\mathbf y^T\mathbf y$
\item the objective function becomes

 $\frac{1}{2} \mathbf{\hat{\pi}}_F^{T}D \mathbf{\hat{\pi}}_{F} + d^T \mathbf{\hat{\pi}}_{F}$
 
\item  Furthermore, the constraints can easily be expressed in the following form: $ A\mathbf{\hat{\pi}}_{F}\leq b $ 
\end{description}

This reformulation fits the quadratic programming with linear constraints framework. Hence, the constrained least squares problem can  be solved using quadratic programming. We used the quadprog package in R for our experiments.

\subsubsection{Least-Absolute-Deviation}

The least absolute deviation problem can be formulated as follows:

\begin{description}
\item min $ \sum{| \mathbf y  - \mathbf X \mathbf{\hat{\pi}}_F |} $

\item s.t.  $ \mathbf{\hat{\pi}}_F>= 0$  and  $\sum\mathbf{\hat{\pi}}_F=1$

\end{description}

This problem can be transformed into a linear programming problem, and solved using standard techniques,  by first defining artificial variables, vector $u$, and rewriting the objective function as:

\begin{description}
\item min $ \sum{u}$

\item s.t.  $u \geq \mathbf y  - \mathbf X \mathbf{\hat{\pi}}_F$
\item        $u \geq -(\mathbf y  - \mathbf X \mathbf{\hat{\pi}}_F)$
\item $\sum\mathbf{\hat{\pi}}_F \geq 1$
\item $-(\sum\mathbf{\hat{\pi}}_F) \geq -1$
\item $ \mathbf{\hat{\pi}}_F>= 0$ 

\end{description}

We used the linprog package  in R to formulate and solve this problem as a linear programming problem. 

\subsubsection{Hellinger Divergence}

The Hellinger divergence problem can be formulated as follows:

\begin{description}
\item min $ 1 - \sum_i{\sqrt{  y_i (\mathbf X \mathbf{\hat{\pi}}_F)_i}} $

\item s.t.  $ \mathbf{\hat{\pi}}_F>= 0$  and  $\sum\mathbf{\hat{\pi}}_F=1$

\end{description}

A reduced gradient optimization approach to solve the dual of the above formulated problem is given in \cite{klafszky_linearly_1989}. Because the regression problem is quite small,  we found out that directly solving this non-linear optimization problem with linear constraints was good enough for  testing purposes.  We used the R package NlcOptim to solve this problem.

\section{Experiments}

We tested the extended versions of the major quantification algorithms using four different datasets: Stanford Sentiment Treebank, CIFAR-10 image data, marketing data presented in \cite{hastie_elements_2009} and 69 Twitter opinion analysis data sets we have created.  Our goal is not to create a definite ranking of the existing methods, but to reveal an embodiment of our unified framework. At the same time, we  empirically show that the multi-class extensions claimed in this paper work well. Although the rankings we get from the experiments provide hints on the viability of the methods, we caution the reader in over-generalizing our results to arrive at a definitive ranking of the quantification methods.

Our R code is made publicly available, and can be used to replicate the experiments.\footnote{https://github.com/aykutfirat/Quantification}

\subsection{Stanford Sentiment Treebank}

We used the training and test split used by \cite{socher_recursive_2013}, of size 8544 and 3311 respectively. We worked with a binary document-term matrix with 3770 features, and 5 classes: very negative, negative, neutral, positive, very positive. 

Instead of using a sophisticated classifier such as Socher's RNTN \cite{socher_recursive_2013}, we used a logistic classifier as our base classifier for quantification. The accuracy of RNTN and logistic regression in this task was 44\%, and 38\% respectively. This was a practical choice for us, because performing cross-validations would take a lot longer if we used RNTN. It is also a test to see if we can get good results with average classifiers. We did use RNTN probability outputs, however, in calculating post-quantification classification adjustments.

In order to test how quantifiers perform in different settings, we considered all potential test-case possibilities with a step proportion size of 0.10. This amounts to ${1/0.1+5-1 \choose 5-1}$=1001 different test cases for 5 classes with step size 0.1 (e.g. (1,0,0,0,0), (0.9,0.1,0,0,0)...(0.2,0.2,0.2,0.2,0.2)..., (0,0,0,0,1))

\subsection{CIFAR-10 dataset}
CIFAR-10 has 60000 colour images in 10 classes (airplane, automobile, bird, cat, deer, dog, frog, horse, ship, truck). We used 55000 of these images as our training data in a bare-bones convolutional neural network (CNN) algorithm that reached about 80\% accuracy. The state of the art CNN algorithms in this data set can reach more than 95\% accuracy with data-augmentation and using ensembles. Our goal, however, was to experiment with an imperfect classifier, as quantification is not needed when we have a perfect classifier (for example in the case of the famous MNIST dataset). 

We used the remaining 5000 images as our test data. We again created all possible test-case possibilities with a step proportion size of 0.10. For practical reasons, however, we randomly sampled 1001 from a total of  92378 cases, and conducted our tests on that sample.

We omitted HDx and VA methods from this experiment, as appropriately converting image pixel data into features is a hard task and would be very important for their ultimate performance.

\subsection{Marketing dataset}

This dataset is an extract from a survey, and consists of 14 demographic attributes with a good mixture of categorical and continuous variables with a lot of missing data. We included this data for experiments as it was used previously by Friedman to test his quantifier. We adopted the same settings Friedman used, namely a classification problem with five occupation classes (Professional/Managerial,  Factory Worker/Laborer/Driver, Homemaker, Student, Retired), 5043 rows and 75 binarized features.

We used a random two thirds of the data as training, and constructed all possible (1001) test combinations with a step size of 0.1 from the remaining test data.

\subsection{Twitter dataset}

Twitter data set consists of 69 different test cases constructed from actual hand-coded tweets. Each test case is about a different topic, and number of classes range from 3 to 12. Data size of each test vary from 700 to 4000, and number of features range from 150 to 4200. We created two types of tests for this data set. 

In the first set of tests, we follow the previous test settings by generating all possible proportions with a variable step size depending on the number of classes. Because the number of test cases can be too many, we sample 20 test proportions randomly for each of the 69 test topics for a total of 1380 tests.

In the second set of tests, we also generate 20 tests from each test topic. This time, however, we start with roughly the training data proportions and generate a random-walk like behavior by sampling from a Dirichlet distribution by using the previous proportion as an input.  This test is intended to simulate the real-life behavior of class proportion changes, where the test set deviates from training gradually. 

In all Twitter tests, we used a training size of 30 per class. The remaining data was partially used for test depending on the test proportions.

\subsection{Results}

We measure the performance of methods using mean absolute deviation (MAD) of the predictions from the true proportions, and also by looking at the post-quantification accuracy. For post-quantification classification we use the post-quantification classification rule by \cite{saerens_adjusting_2002}: 

$\hat{P}_{kF}(x) =\frac{\frac{\hat{\pi}_{kF}}{\pi_{kT}}\hat{P}_{kT}(x)}{\sum_{l=1}^K\frac{\hat{\pi}_{lF}}{\pi_{lT}}\hat{P}_{lT}(x)}  $.

We show the average MAD results in Table 2, and average post-quantification classification accuracy results in Table 3. Results are the averages of scores over all experiments per data set. Accuracy is simply the number of correct classifications divided by the total number of classifications. In both tables, ``Naive'' shows the results of naive quantification, a base-line for performance. In Table 3, ``Truth'' row shows the accuracy obtainable if we knew the test proportions precisely, an upper-bound for what is achievable.

In Figure 1, we show the best performing quantification methods for the Stanford, Marketing and CIFAR-10 data sets along with the actual test proportions, and naive quantification. 1001 experiments are stacked together in one graph for a finer look at the quantification performance. Each horizontal line corresponds to an experiment, and each color in that line is category whose length is its proportion. The punctuated nature of the graphs is due to the ordering of categories for easier comparability.

In Figure 2, the post-quantification accuracy densities are displayed for the best performing methods. The density graphs reveal that quantification benefits are higher when the gap between test and training distributions are higher, and the base-classifier does not have very high accuracy like the CIFAR-10 case.

\begin{table}
\begin{center}
\begin{tabular}{|lrrrr|}
 \hline
 Quantifier & \multicolumn{4}{c|}{\% MAD}    \\ 

   \bf  & Stan. & C-10 & Market & Twitter   \\ 
  \hline
Prob & \bf 4.93 & 0.39 &  1.67 & {\bf  9.13}/6.00 \\
MM& 5.11 & 0.39 & \bf1.62 &   9.48/\bf 5.47 \\
AC &  5.53 & 0.40 & 2.00 & 9.60/5.88\\ 
FM &  6.11 & 0.43 & 1.85 & 10.10/5.96 \\ 
MS & 6.29 & 0.40 & 1.78 & 11.87/5.99\\
HDy &  8.34 & \bf 0.36 & 1.86 & 11.81/6.10  \\
VA &  10.66 & - & 2.17 & 11.22/6.16\\ 
HDx & 10.99 & - & 1.98 & 13.02/6.69 \\

  \hline
Naive &  14.71 & 3.65 & 10.35 & 14.46/7.07\\
\hline
\end{tabular}
\end{center}
\caption{Average MAD scores for experiments. Naive shows quantification based on averaging probability outputs of the classifier. Bold numbers show the best result within that test set. }
\end{table}

\begin{table}
\begin{center}
\begin{tabular}{|lrrrr|}
 \hline
 Quantifier & \multicolumn{4}{c|}{\% Post-Quantification Accuracy}    \\ 
  \bf  & Stan. & C-10 & Market & Twitter   \\ 
  \hline
Truth &  63.6 & 87.3  &  82.2 &  72.6/58.7\\
  \hline
Prob & \bf 60.7 & \bf 87.0  & 81.7  & {\bf 64.0}/54.4 \\
MM& 60.1 &\bf 87.0  &\bf 81.8  & 63.7/55.1 \\
AC &  59.7 & 86.9 &81.6  & 63.5/54.4\\ 
FM &  58.7 & 86.9 & 81.6  & 62.9/54.4 \\ 
MS & 58.7 & \bf 87.0 &81.7 & 61.0/\bf55.2\\
HDy &  52.5 &\bf 87.0  & 81.7   & 61.3/\bf 55.2  \\
VA & 49.8&  - & 81.5 & 61.4/54.9\\ 
HDx & 48.2 & - & 81.6  & 59.5/55.0 \\
  \hline
Naive &  37.7 &  76.8 &  67.4 &  53.8/54.1\\
\hline
\end{tabular}
\end{center}
\caption{Average Post-Quantification  Accuracy. Naive shows the average accuracy we get without any quantification adjustment, while Truth shows what is achievable if we knew the actual test set proportions. }
\end{table}

\subsubsection{Post-Quantification Classification Performance}
One important benefit of quantification is improved classification performance. In Figure 4, we show how different quantification approaches increase classification accuracy by fitting a regression curve for the classification accuracies as they vary with respect to MAD of test proportions from training proportions. Naive classifier performance gets worse as the MAD gets larger. The best methods show fast increasing classifier accuracy as test and training proportions diverge.
\begin{figure}

\includegraphics[width=8cm, height=8cm]{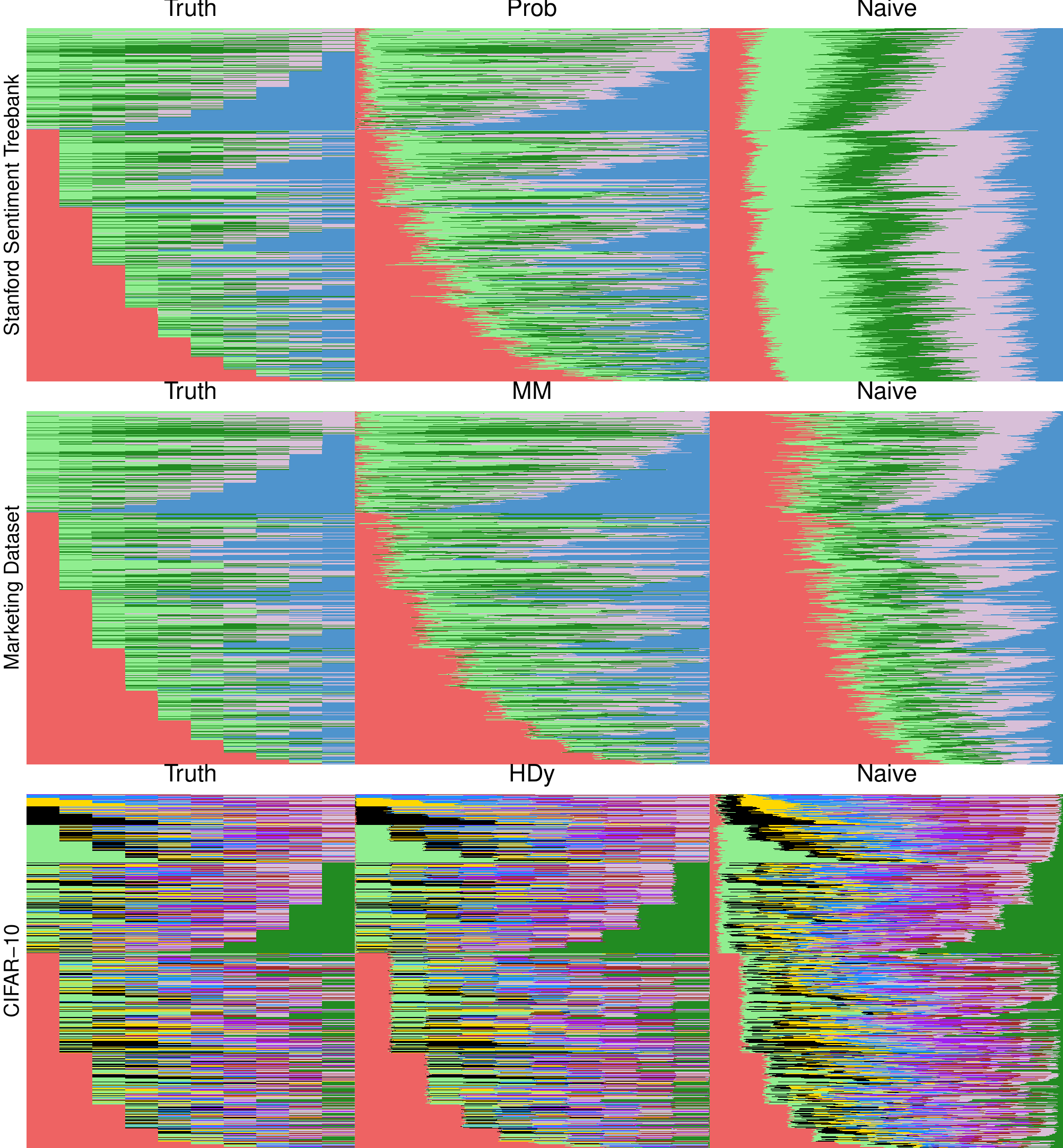}
\caption{Graphical experiment results in 1001 tests in Stanford Sentiment Treebank, Marketing Dataset and CIFAR-10 image set. Each horizontal line corresponds to one of the 1001 experiments, each color is a category and its horizontal length corresponds to its proportion. Best performer for each dataset is compared to Naive quantification, and Truth. As evident from the graphs, the quantification task is hardest in Stanford data, and easiest in CIFAR-10.}

\end{figure}

\begin{figure}

\includegraphics[width=8cm, height=8cm]{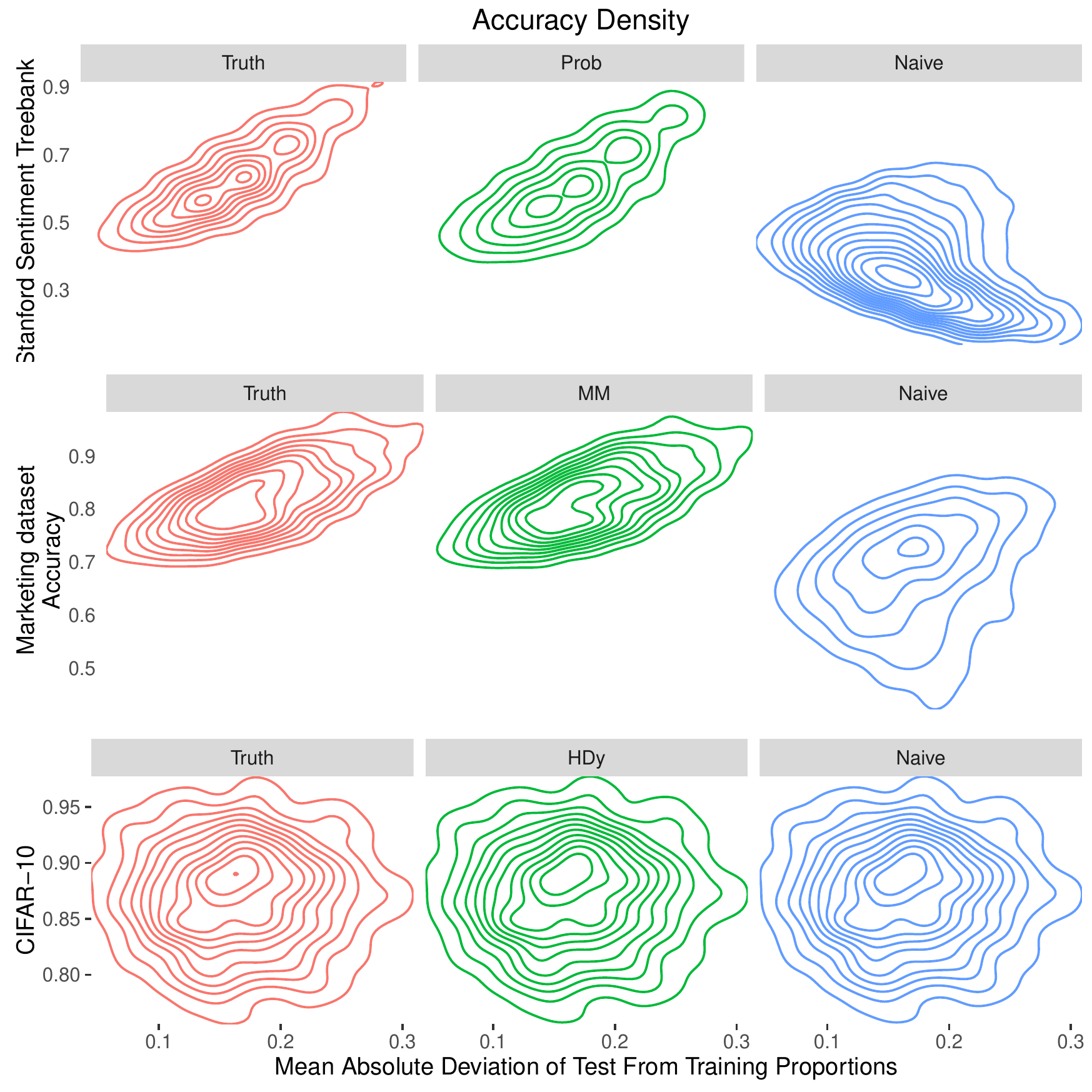}
\caption{Accuracy Density in Stanford Sentiment Treebank, Marketing Dataset and CIFAR-10 image set. The improvement in accuracy, and the alignment of quantification with Truth, is clearly visible in harder domains such as the Stanford Sentiment Treebank and the Marketing dataset. The accuracy densities look very similar in an easy quantification data set such as the CIFAR-10.}

\end{figure}

\begin{figure}

\includegraphics[width=8cm, height=8cm]{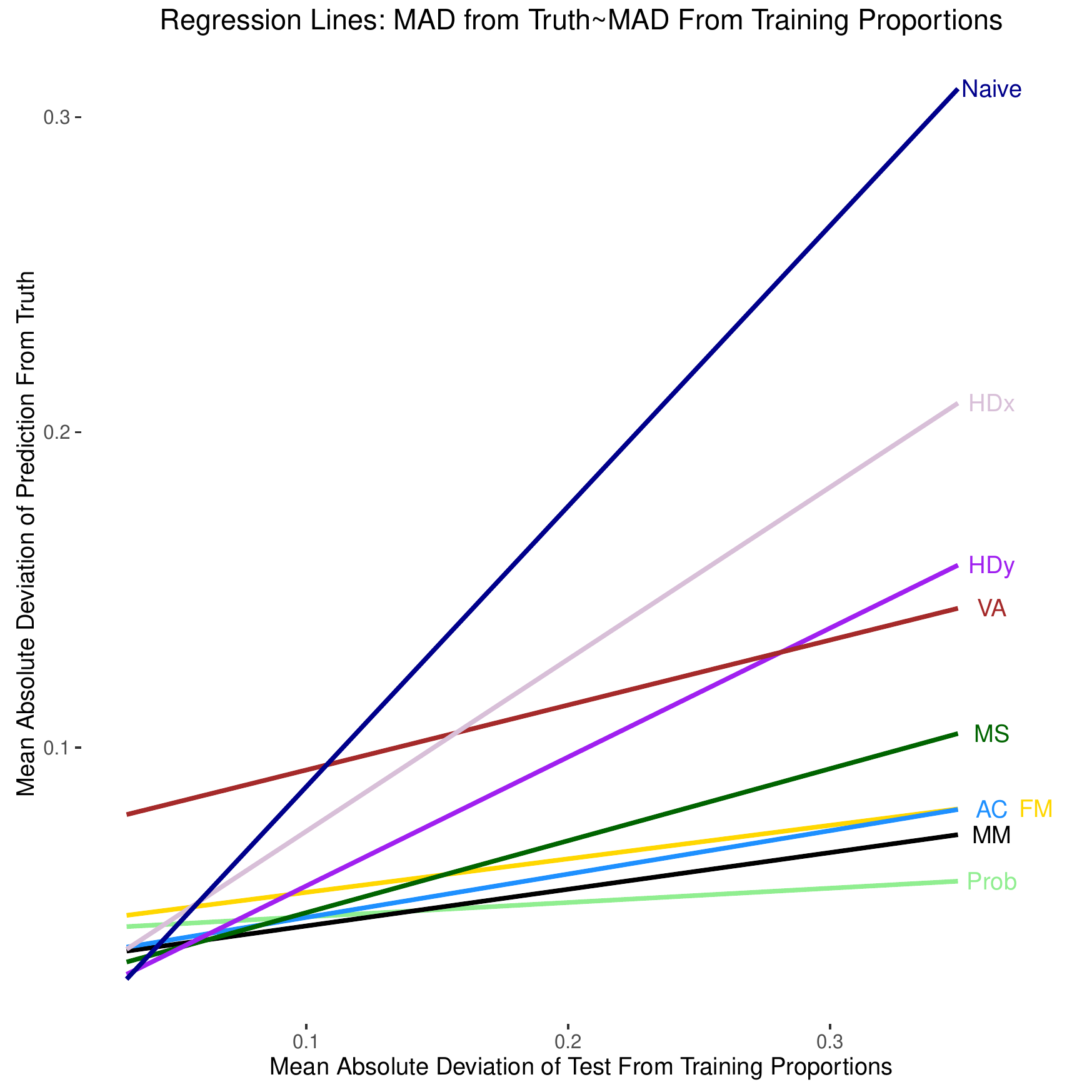}
\caption{Regression lines showing how each algorithm behaves as the test proportions deviate from the training proportions in the Stanford Sentiment Treebank tests}
\end{figure}

\begin{figure}

\includegraphics[width=8cm, height=8cm]{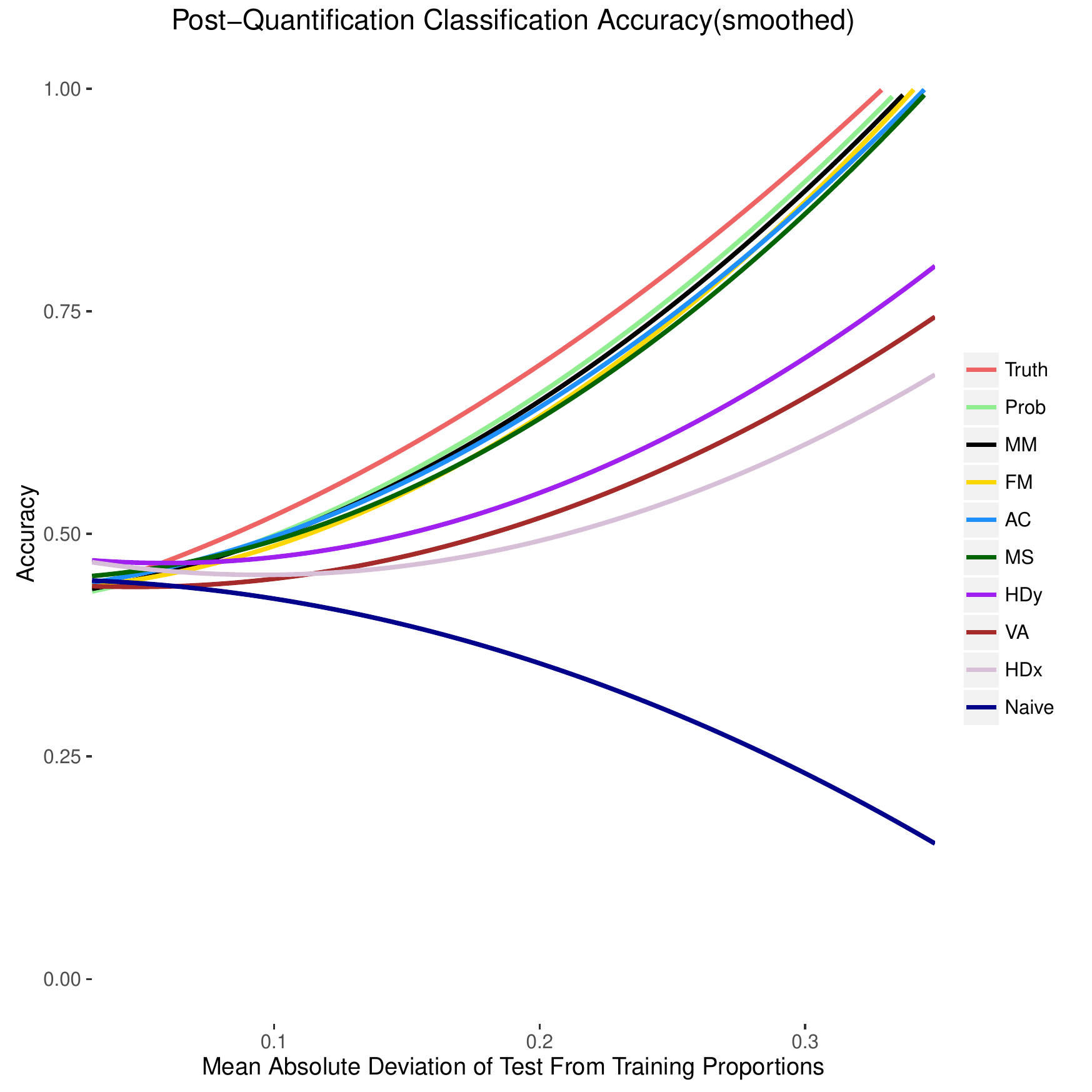}
\caption{Regression curves showing how the post-quantification accuracy changes as the test proportions deviate from the training proportions in the Stanford Sentiment Treebank tests}
\end{figure}
\subsubsection{Performance Sensitivity to Training Proportions}
We show in Figure 3, how the performance of quantification methods vary as test proportions deviate from the training proportions. We fit a regression line for each quantification method by using MAD of test proportions from the training proportions as the independent variable, and MAD of the predicted test proportions from the true test proportions as the dependent variable. The best quantifiers are expected to be insensitive to the training proportions. The Prob and MM variations, display remarkable insensitivity to training proportions, while the Hellinger divergence based methods have higher sensitivity.

\section {Conclusion}
We presented a constrained regression framework for unifying major quantification approaches. Under this framework, quantification approaches primarily differ on their choice of a feature transformation function that is expected to remain near constant within classes in training and test data, and be sufficiently different across classes. Using mathematical programming, we also offered solution methodologies for least-squares, least-absolute deviation, and Hellinger divergence loss functions for the constrained regression problem. Furthermore, we extended the binary-only quantification approaches to multi-class settings, and presented the results of our experiments with four different data sets to verify that our multi-class extensions work as expected in practice. Our code is made available for replication and further experimentation.

\bibliography{quantification}
\bibliographystyle{emnlp2016}

\end{document}